\documentclass[11pt]{article}
\usepackage{acl2016}
\usepackage{times}
\usepackage{latexsym}

\usepackage{graphicx}
\usepackage{algorithm,algorithmic}

\usepackage{multirow}
\usepackage{graphicx}
\usepackage{subcaption}

\usepackage{amsmath}
\usepackage{amssymb}
\usepackage{color,soul}
\usepackage{url}

\usepackage{threeparttable}

\aclfinalcopy 

\title{Learning when to trust distant supervision: An application to low-resource POS tagging using cross-lingual projection}

\author{Meng Fang \and Trevor Cohn \\
         Department of Computing and Information Systems, The University of Melbourne\\
         \url{meng.fang@unimelb.edu.au}, \url{t.cohn@unimelb.edu.au}}

\date{}

\begin{document}

\maketitle

\begin{abstract}
Cross lingual projection of linguistic annotation suffers from many sources of bias and noise, leading to unreliable annotations that cannot be used directly. In this paper, we introduce a novel approach to sequence tagging that learns to correct the errors from cross-lingual projection using an explicit debiasing layer. This is framed as joint learning over two corpora, one tagged with gold standard and the other with projected tags. We evaluated with only 1,000 tokens tagged with gold standard tags, along with more plentiful parallel data. Our system equals or exceeds the state-of-the-art on eight simulated low-resource settings, as well as two real low-resource languages, Malagasy and Kinyarwanda.
\end{abstract}

\section{Introduction}
Part-of-speech (POS) tagging is a critical task for natural language processing (NLP) applications, providing lexical syntactic information. 
Automatic POS tagging has been extremely successful for many rich resource languages through the use of supervised learning over large training corpora~\cite{mccallum2000maximum,lafferty2001conditional,ammar2016many}. However, learning POS taggers for low-resource languages from small amounts of annotated data is very challenging~\cite{garrette2013learning,duong2014can}. For such problems, distant supervision via heuristic methods can provide cheap but inaccurately labelled data~\cite{mintz2009distant,takamatsu2012reducing,ritter2013modeling,plank2014adapting}. A compromise, considered here, is to use a mixture of both resources: a small collection of clean annotated data and noisy ``distant'' data.

A popular method for distant supervision is to use parallel data between a low-resource language and a rich-resource language. Although annotated data in low-resource languages is difficult to obtain, bilingual resources are more plentiful. For example parallel translations into English are often available, in the form of news reports, novels or the Bible. Parallel data allows annotation from a high-resource language to be projected across alignments to the low-resource language,  which has been shown to be effective for several language processing tasks including POS tagging~\cite{yarowsky2001inducing,das2011unsupervised},  named entity recognition~\cite{wang2013cross} and dependency parsing~\cite{mcdonald2013universal}.

Although cross-lingual POS projection is popular it has several problems, including errors from poor word alignments and cross-lingual syntactic divergence~\cite{tackstrom2013token,das2011unsupervised}. Previous work has proposed heuristics or constraints to clean the projected tag before or during learning. 
In contrast, we consider compensating for these problems explicitly, by learning a bias transformation to encode the mapping between `clean' tags and the kinds of tags produced from projection.

We propose a new neural network model for sequence tagging in a low-resource language, suitable for training with both a tiny gold standard annotated corpus, as well as distant supervision using cross-lingual tag projection. Our model uses a bidirectional Long Short-Term Memory (BiLSTM), which produces two types of output: gold tags generated directly from the hidden states of a neural network, and uncertain projected tags generated after applying a further linear transformation. This transformation, encodes the mapping between the projected tags from the high-resource language, and the gold tags in the target low-resource language, and learns when and how much to trust the projected data. For example, for languages without determiners, the model can learn to map projected determiner tags to nouns, or if verbs are often poorly aligned, the model can learn to effectively ignore the projected verb tag, by associating all tags with verbs. Our model is trained jointly on gold and distant projected annotations, and can be trained end-to-end with backpropagation. 

Our approach captures the relations among tokens, noisy projected POS tags and ground truth POS tags. Our work differs in the use of projection, in that we explicitly model the transformation between tagsets as part of a more expressive deep learning neural network. We make three main contributions. First, we study the noise of projected data in word alignments and describe it with an additional transformation layer in the model. Second, we integrate the model into a deep neural network and jointly train the model on both annotated and projected data to make the model learn from better supervision. Finally, evaluating on eight simulated and two real-world low-resource languages, experimental results demonstrate that our approach uniformly equals or exceeds existing methods on simulated languages, and achieves 86.7\% accuracy for Malagasy and 82.6\% on Kinyarwanda, exceeding the state-of-the-art results of \newcite{duong2014can}.

\section{Related Work}

For most natural language processing tasks, the conventional approach to developing a system is to use supervised learning algorithms trained on a set of annotated data. However, this approach is inappropriate for low-resource languages due to the lack of annotated data. An alternative approach is to harness different source of information aside from annotated text. Knowledge-bases such as dictionaries are one such source, which can be used to inform or constrain models, such as limiting the search space for POS tagging~\cite{banko2004part,goldberg2008can,li2012wiki}.

Parallel bilingual corpora provide another important source of information. These corpora are often plentiful even for many low-resource languages in the form of multilingual government documents, book translations, multilingual websites, etc. Word alignments can provide a bridge to project information from a resource-rich source language to a resource-poor target language. For example, parallel data has been used for named entity recognition~\cite{wang2013cross} based on the observation that named entities are most often preserved in translation and also in syntactic tasks such as POS tagging~\cite{yarowsky2001inducing,das2011unsupervised} and dependency parsing~\cite{mcdonald2013universal}. Clues from related languages can also compensate for the lack of annotated data, as we expect there to be information shared between closely related languages in terms of the lexical items, morphology and syntactic structure. 
Some successful applications using language relatedness information are dependency parsing~\cite{mcdonald2011multi} and POS tagging~\cite{hana2004resource}.
However, these approaches are limited to closely related languages such as Czech and Russian, or Telugu and Kannada, and it is unclear whether these techniques will work well in situations where parallel data only exists for less-related languages, as is often the case in practice.

To summarize, for all these mentioned tasks, lexical resources are valuable sources of knowledge, but are also costly to build. Language relatedness information is applicable for closely related languages, but it is often the case that a given low-resource language does not have a closely-related, resource-rich language. Parallel data therefore appears to be the most realistic additional source of information for developing NLP systems for low-resource languages~\cite{yarowsky2001inducing,duong2014can,guo2015cross}, and here we primarily investigate methods to exploit parallel texts. 

\newcite{yarowsky2001inducing} pioneered the use of parallel data for projecting POS tag information from a resource-rich language to a resource-poor language. \newcite{duong2014can} proposed an approach using a maximum entropy classifier trained on 1000 tagged tokens, and used  projected tags as auxiliary outputs. \newcite{das2011unsupervised} used parallel data and exploited graph-based label propagation to expand the coverage of labelled tokens. Our work is closest to~\newcite{duong2014can}, and we share the same evaluation setting, which we believe is well suited to the low-resource applications. Our approach differs from theirs in two ways: first we propose a deep learning model based on a long short-term memory recurrent structure versus their maximum entropy classifier, and secondly we model the projection tag explicitly as a biased variant of the classification output, while they attempt to capture the correlations between tagsets only implicitly through a joint feature set over both tags. We believe that our work is the first to explicitly model the bias affecting cross-lingual projected annotations, thereby allowing this rich data resource to be better exploited for learning NLP models in low-resource languages. 

\section{Framework}
In this work, we consider the POS tagging problem for a low-resource language using both the gold annotated and distant projected corpora. For a low-resource language, we assume two sets of data. First, there is a small conventional corpus for the low-resource language, annotated with gold tags. Second, there is a parallel corpus between the language and English, where we can reliably tag the English side and project these annotations across the word alignments. Then based on the annotated and the projected data, we learn a deep neural model for the POS tagging. The goal of learning here is to improve the POS tagging accuracy on the low-resource language.

\subsection{POS projection via word alignments}
Parallel data is often available for low-resource languages. For example, for Malagasy we can obtain bilingual documents with English directly from the web. This provides ample opportunity for projecting annotations from English into the low-resource language. Although the POS tags can be projected, given sentence and word-alignments, direct projection has several issues and results in noisy, biased and often unreliable annotations~\cite{yarowsky2001inducing,duong2014can}. One source of error are the word alignments. These errors arise from words in the source language that are not aligned to any words in the target language, which might be due to them not being translated well enough, errors in alignments, or translation phenomena that do not fit the assumptions underlying the word based alignment models (e.g., many-to-many translations cannot be captured). 

\begin{figure}
    \centering
    \includegraphics[width=0.5\textwidth]{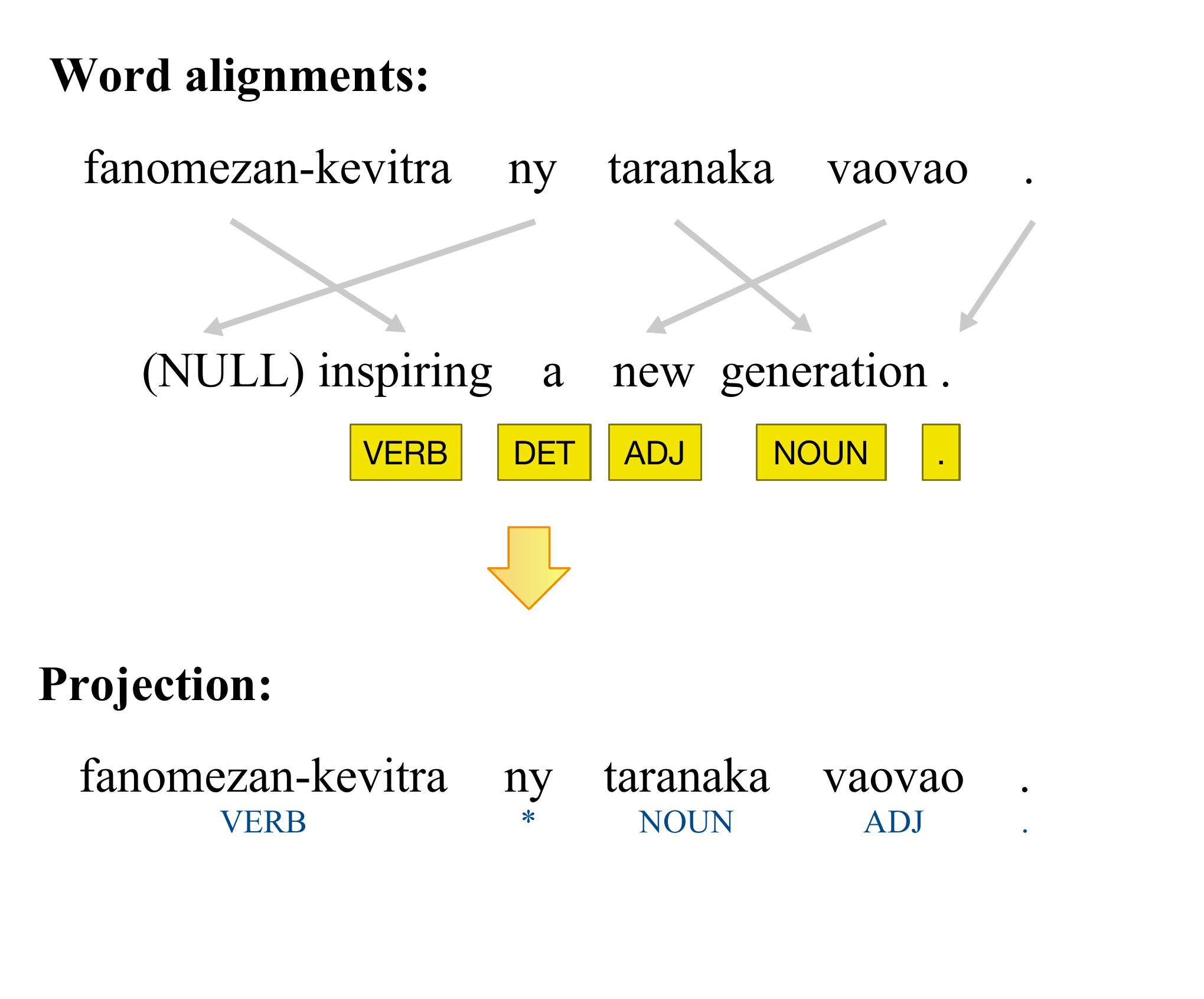}
     \vspace{-28pt}
    \caption{An example of POS projection via word alignments. * indicates unknown POS tag, which we treat as having a tag distribution over all tokens in the source sentence (in the example, a uniform mix of VERB, DET, ADJ, NOUN and `.').}
    \label{fig-align}
\end{figure}

An example of POS projection via word alignments between Malagasy and English is shown in Figure~\ref{fig-align}. A word in Malagasy is connected to a word in English or the NULL word. Thus there exist words in the target language which are not aligned to a word in the source language, for example \emph{ny} in Figure~\ref{fig-align}. Previous work has either used the majority projected POS tag for a token or used a default value to represent the token~\cite{duong2014can,tackstrom2013token}. Another problem are errors in the projected tags: for example, in this sentence, \emph{fanomezan-kevitra} is labelled as VERB incorrectly, but should be NOUN, a consequence of a non-literal translation. 
 
We now turn to the labelling of the projected data. For the parallel data, we consider each token in the low-resource language. Where this token is aligned to a single token in English, we assign the tag for that English token. For tokens that are aligned to many English words or none at all (NULL), we assign a distribution over tags according to the tag frequency distribution over the whole English sentence.

A natural question is whether this projected labelling might be suitable for use directly in supervised learning of a POS tagger. To test this, we compare training a bidirectional Long Short-Term Memory (BiLSTM) tagger on this data, a small 1000 token dataset with  gold-standard tags, and the union of the two.\footnote{See \S\ref{sec:tagger_with_noise} for the model details, and \S\ref{sec:data} for a description of the datasets and evaluation.} Evaluating the tagging accuracy against gold standard tags, we observe in Tables~\ref{tab-8} and~\ref{tab-mlg-kin} (top section, rows labelled BiLSTM) that the use of the gold-standard (Annotated) data is considerably superior to training on the directly Projected data, despite the smaller amount of Annotated data, while using the union of the two datasets results in mild improvements in a few languages, but worsens performance for others.

These sobering results raise the  question of how we might use the bilingual resources in a more effective manner than direct projection. Clearly projections contain useful information, as the tagging accuracy is well above chance. However, they are riddled with noise and biases, which need to be accounted for to improve performance.

\subsection{BiLSTM with bias layer} \label{sec:tagger_with_noise}
\begin{figure*}[t]
    \centering
    \includegraphics[width=0.8\textwidth]{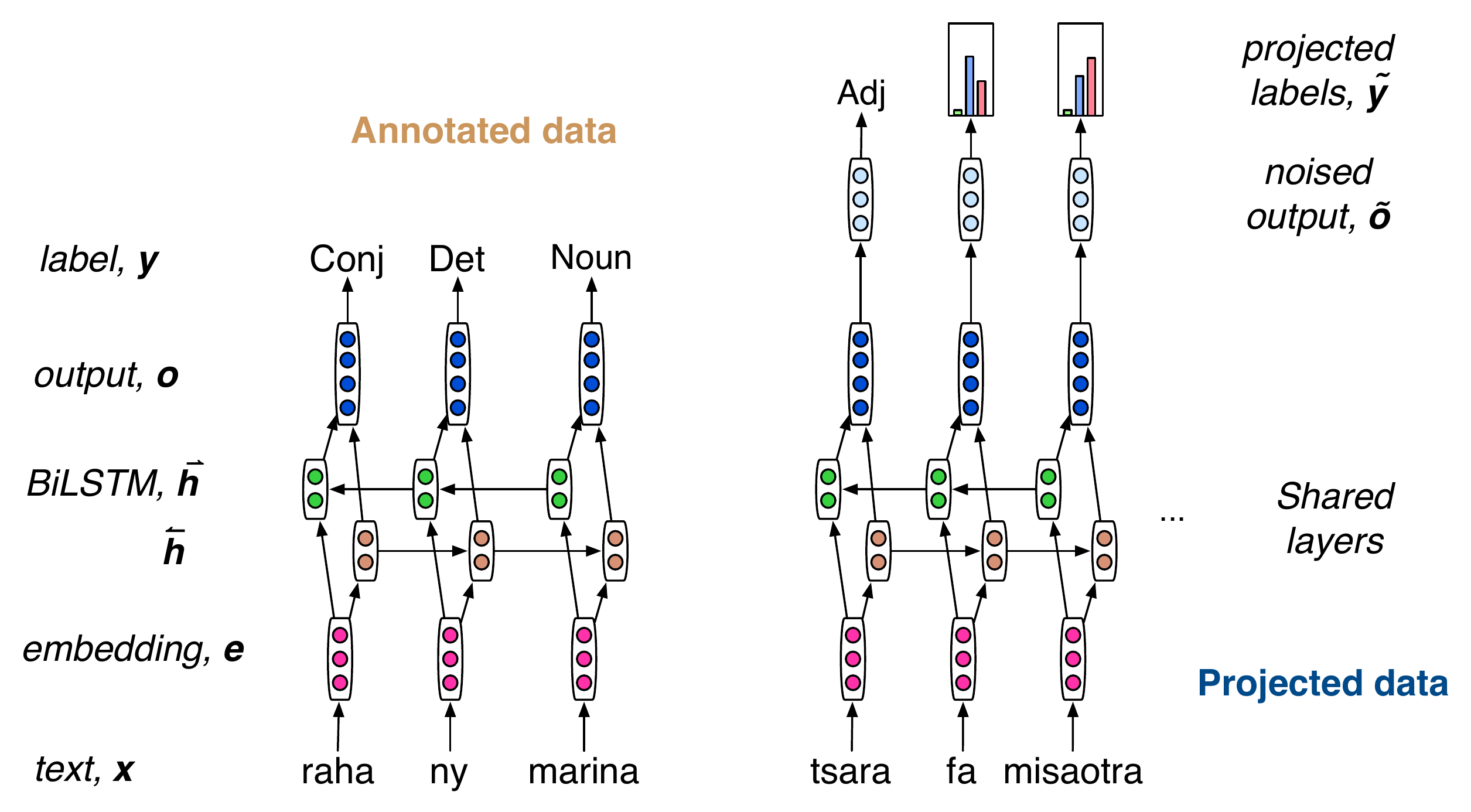}
    \caption{Illustration of the model architecture, which uses a bidirectional LSTM recurrent network, with a tag classification output. The left part illustrates the supervised training scenario and test setting, where each word $x$ is assigned a tag $y$; the right part shows the projection training setting, with a bias layer, where the supervision is either a projected label or label distribution (used for NULL aligned words).}
    \label{fig-2models}
\end{figure*}

To address this problem, we propose a model that jointly models the clean annotated data and the projected data. For this we use a bidirectional LSTM tagger, as illustrated on the left in Figure~\ref{fig-2models}, although other classifiers could be easily used in its place. The BiLSTM  offers access to both the left and right lexical contexts around a given word~\cite{graves2013speech}, which are likely be of considerable use in POS tagging where context of central importance.

Let $x_t$ indicate a word in a sentence and $y_t$ indicate its corresponding POS tag, and $K$ denotes the size of the tagset.\footnote{We use the universal tagset from~\newcite{petrov2011universal}, enabling easier comparison with related work, although this is not a requirement of our work.} The recurrent layer is designed to store contextual information, while the values in the hidden and output layers are computed as follows:
\begin{align}
\overrightarrow{h}_t &=\mathtt{lstm}(\overrightarrow{h}_{t-1},x_t) \nonumber\\
\overleftarrow{h}_t & =\mathtt{lstm}(\overleftarrow{h}_{t+1},x_t) \nonumber\\
o_t & =\text{softmax}(W_{\rightarrow} \overrightarrow{h}_t + W_{\leftarrow} \overleftarrow{h}_t + b) \label{eq:ot} \\
y_t & \sim \text{Multinomial}(o_t) \, . \nonumber
\end{align}
This supervised model is trained on annotated gold data in the standard manner using a cross-entropy objective with stochastic gradient descent through the use of gradient backpropagation.

The projected data, however, needs to be treated differently to the annotated data: the tagging is often uncertain, as tokens may have been aligned to words with different parts of speech, or multiply aligned, or left as an unaligned word. These tags are not to be trusted in the same way as the gold annotated data. Our work accounts for bias explicitly in the training objective, by modelling the correspondence between the true tags and the errorful projected tags. The projected data consists of pairs, $(x_t, \tilde{y})$, where $\tilde{y}$ denotes the projected POS tag or tag distribution. In this setting, we assume that the true label, $y_t$, is latent variable and both $\tilde{y}$ and $y$ are $K$-dimensional binary random variables: $\tilde{y}_t$ is a vector representation of a projected tag, and $y_t$ is a one-hot representation of a gold tag. 

We augment the deep neural network model to include a bias transformation such that its prediction matches the distribution of the projected tags, as follows:
\begin{equation}
p(\tilde{Y}_t=j|x_t, \theta, A) = \operatorname{softmax}\left(\sum_{i}a_{i,j}o_{t,i}\right) \, ,
\label{eq-combined}
\end{equation}
where $o_{t,i} = p(Y_t = i|x_t,\theta)$ is the probability of tag $i$ in position $t$ according to (\ref{eq:ot}). This equation is parameterized by a $K \times K$ matrix $A$.\footnote{Our approach also supports mismatching tagsets, in which case $A$ would be rectangular with dimensions based on the sizes of the two tag sets.} Each cell $a_{i,j}$ denotes the confusion score between classes $i$ and $j$, with negative values quashing the correspondance, and positive values rewarding a pairing; in the situations where the projected tags closely match the supervised tagging, we expect that $A \propto I$.

Joint modelling of the gold supervision and projected data gives rise to a training objective combining two cross-entropy terms, 
\begin{align*} 
\mathcal{L}( \theta, A) = 
&-\frac{1}{|T^{p}|} 
\sum_{t \in T^{p}} \langle 
\tilde{y}_{t} ,
\log \operatorname{softmax}\left( A o_t \right) \rangle  \\
&- \frac{1}{|T^{t}|} 
\sum_{t \in T^{t}} \langle 
y_{t} ,
\log o_t \rangle
\, ,
\end{align*}
where $T^{p}$ indexes all the token positions in the projected dataset, and $T^{t}$ does similarly for the annotated training set. 

We illustrate the combined model in Figure ~\ref{fig-2models}, showing on the left the gold supervised model and on the right the distant supervised components. The distant model builds on the base part by feeding the output through a bias layer, which is finally used in a softmax to produce the biased output layer. The matrix $A$ parameterizes the final layer, to adjust the tag probabilities from the supervised model into a distribution that better matches the projected POS tags. However, the ultimate goal is to predict the POS tag $y_t$. Consider the training effect of the projected POS tags: when performing error backpropagation, the cross-entropy error signal must pass through the tag transformation linking $\tilde{o}$ with $o$, which can be seen as a debiasing step, after which the cleaned error signal can be further backpropagated to the rest of the model. Provided there are consistent patterns of errors in the projection output, this technique can readily model these sources of variation with a tiny handful of parameters, and thus greatly improve the utility of this form of distant supervision.

Directly training the whole deep neural network with random initialization is impractical, because without a good estimate for the $A$ matrix, the errors from the projected tags may misdirect training result in a poor local optima. For this reason the training process contains two stages. In the first stage we use the clean annotated data to pretrain the network. In the second stage we jointly use both projected and annotated data to continue training the model.

\section{Experiments}

\begin{table*}[t]
\centering
\begin{tabular}{l rc rc rc rc rc rc rc rc | c}
\hline ~ & da & nl & de & el & it & pt & es & sv & Average \\ \hline
BiLSTM Annotated & 89.3 & 87.4  & 89.5 & 88.1  & 85.9 & 89.5  & 90.6 & 84.7 & 88.1 \\
BiLSTM Projected & 64.4 & 81.9  & 81.3 & 78.9  & 80.1 & 81.9  & 81.2 & 74.9  & 78.0 \\
BiLSTM Ann+Proj & 85.4 & 88.9  & 90.2 & 84.2  & 86.1 & 88.2  & 91.3 & 83.6 & 87.2\\
\hline
MaxEnt Supervised & 90.1 & 84.6 & 89.6  & 88.2 &  81.4  & 87.6  & 88.9  & 85.4 & 86.9\\
\newcite{duong2014can} & 92.1 & 91.1  & 92.5 & 92.1  & 89.9 & 92.5  & 91.6 & 88.7  & 91.3\\
BiLSTM Debias & 92.3 & 91.7  & 92.5 & 92.8  & 90.2 & 92.9  & 92.4 & 89.1 & 91.7\\
\hline
\end{tabular}
\caption{The POS tagging accuracy for various models in eight languages: Danish (da), Dutch (nl), German (de), Greek (el), Italian (it), Portuguese (pt), Spanish (es), Swedish (sv). The top results of the second part are taken from~\newcite{duong2014can}, evaluated on the same data split.}
\label{tab-8}
\end{table*}

We evaluate our algorithm using two kinds of experimental setups, simulation experiments and real-world experiments.
For the simulation experiments, we use the following eight European languages: Danish (da), Dutch (nl), German (de), Greek (el), Italian (it), Portuguese (pt), Spanish (es), Swedish (sv). These languages are obviously not low-resource languages, however we can use this data to simulate the low-resource setting by only using a small 1,000 tokens of the gold annotations for training. This evaluation technique is widely used in previous work, and allows us to compare our results with prior state-of-the-art algorithms. For the real-world experiments, we use the following two low-resource languages: Malagasy, an Austronesian language spoken in Madagascar, and Kinyarwanda, a Niger-Congo language spoken in Rwanda.  

\subsection{Evaluation Corpora}
\label{sec:data}

\subsubsection{Parallel data}
For the simulation experiments, we use the Europarl v7 corpus, with English as the source language and each of languages as the target language. There are an average of 1.85 million parallel sentences for each of the eight language pairs. 
For the real-world experiments, the parallel data is smaller and generally of a lower quality. 
For Malagasy, we use a web-sourced collection of parallel texts.\footnote{\url{http://www.cs.cmu.edu/~ark/global-voices}} The parallel data of Malagasy has 100k sentences and 1,231k tokens. For Kinyarwanda, we obtained parallel texts from ARL MURI project.\footnote{The dataset was provided directly by Noah Smith.}, constituting 11k sentences and 52k tokens.

\subsubsection{POS projection}
We use GIZA++ to induce word alignments on the parallel data~\cite{och2003systematic}, using IBM model 3~\cite{brown1993mathematics}. Following prior work~\cite{duong2014can}, we retain only one-to-one alignments. Using all alignments (i.e., many-to-one and one-to-many), would result in many more POS-tagged tokens, but also bring considerable additional noise. For example, the English \emph{laws} (NNS) aligned to French \emph{les} (DT) \emph{lois} (NNS) would end up incorrectly tagging the French determiner \emph{les} as a noun (NNS). We use the Stanford POS tagger~\cite{toutanova2003feature} to tag the English side of the parallel data and then project the labels to the target side. As we show in the following section, and confirmed in many studies~\cite{tackstrom2013token,das2011unsupervised}, the directly projected labels have many errors and therefore it is unwise to use the tags directly. We further filter the corpus using the approach of \newcite{yarowsky2001inducing} which selects sentences with the highest sentence alignment scores from IBM model 3. For the European languages, we retain 200k sentences for each language, while for the low-resource languages, we use all the parallel data. 

\begin{figure*}
    \includegraphics[width=\textwidth]{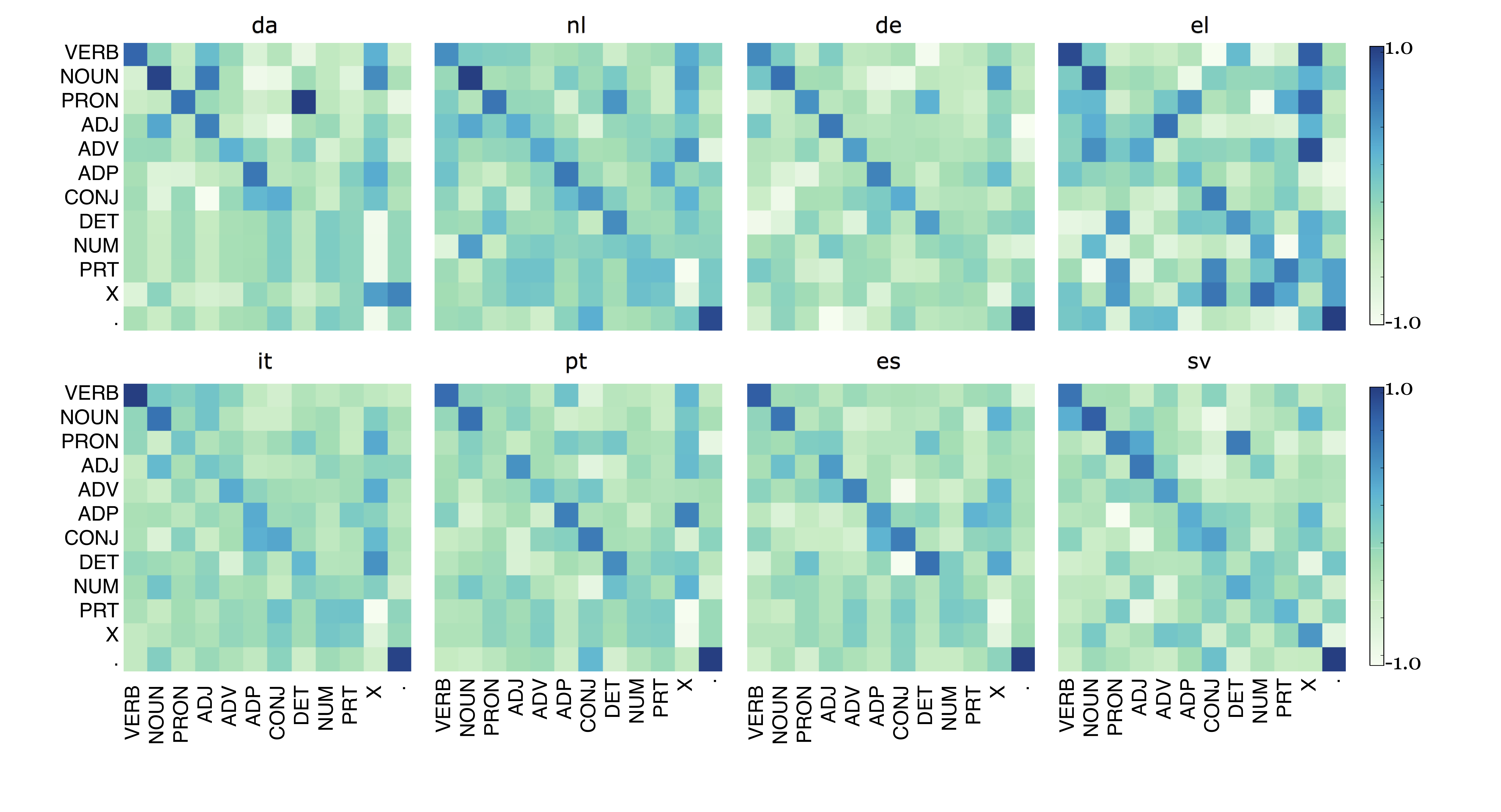}
    \caption{Bias transformation matrix $A$ between POS tags and projection outputs, shown respectively as columns and rows for the eight languages.}
    \label{fig-8}
\end{figure*}

\subsubsection{Annotated data}
Gold annotated data is expensive and difficult to obtain, and thus we assume that only a small annotated dataset is available. For the simulation experiments, annotated data is obtained from the CoNLL-X shared tasks \cite{buchholz2006conll}. To simulate the low-resource setting, we take the first 1,000 tagged tokens for training and the remaining data is split equally between development and testing sets, following \newcite{duong2014can}. For the real-world experiments, we use the Malagasy and Kinyarwanda data from \newcite{garrette2013learning}, who showed that a small annotated dataset could be collected very cheaply, requiring less than 2 hours of non-expert time to tag 1,000 tokens. This constitutes a reasonable demand for cheap portability to other low-resource languages. We use the datasets from \newcite{garrette2013learning}, constituting annotated datasets of 383 sentences and 5,294 tokens in Malagasy and 196 sentences and 4,882 tokens for Kinyarwanda. We use 1,000 tokens as training set and the rest is used for testing for each language.

\subsection{Setup and baselines} 
We compare our algorithm with several baselines, including the state-of-the-art algorithm from \newcite{duong2014can}, a two-output maxent model, their reported baseline method of a supervised maximum entropy model trained on the annotated data, and our BiLSTM POS tagger trained directly from the annotated and/or projected data (denoted BiLSTM Annotated, Projected and Ann+Proj for the model trained on the union of the two datasets). For the real low-resource languages, we also compare our algorithm with~\newcite{garrette2013real}, who reported good results on the two low-resource languages. Our implementation is based on the \texttt{cnn} toolkit.\footnote{\url{https://github.com/clab/cnn}} In all cases, the BiLSTM models use 128 dimensional word embeddings and 128 dimensional hidden layers. We set the learning rate to 1.0 and use stochastic gradient descent model to learn the parameters.

We evaluate all algorithms on the gold testing sets, evaluating in terms of tagging accuracy. Following standard practice in POS tagging, we report results using per-token accuracy (i.e., the fraction of predicted tags that exactly match the gold standard tags). Note that for all our experiments, we work with the universal POS tags and accordingly accuracy is measured against the gold tags after automatic mapping into the universal tagset.

\subsection{Results}

\begin{figure*}[t]
\centering
\includegraphics[width=0.8\textwidth]{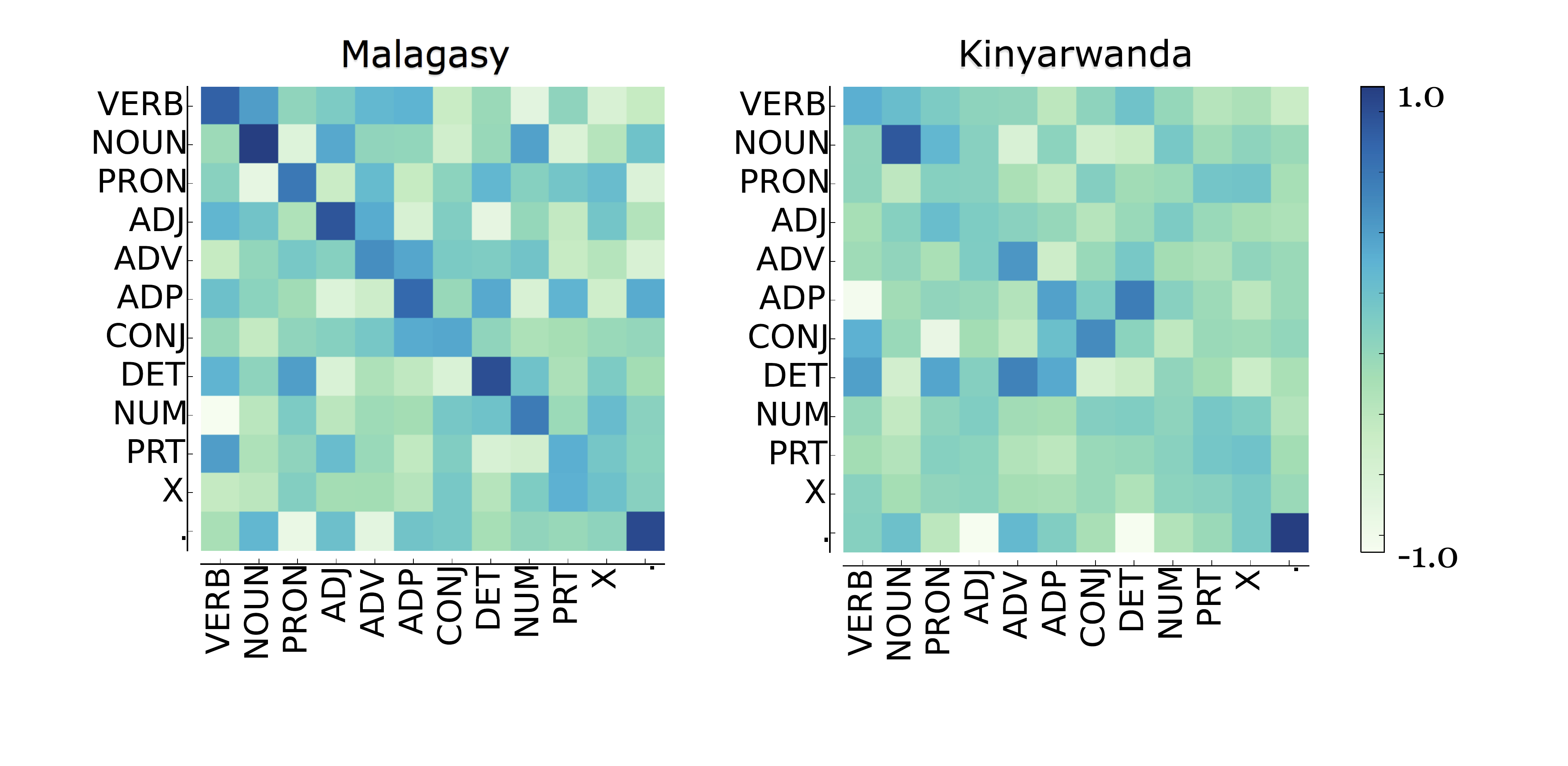}
\caption{Bias transformation matrix $A$ between POS tags and projected outputs, shown respectively as columns and rows for the two low-resource languages.}
\label{fig-mlg-kin-nl}
\end{figure*}

First, we present the results for the 8 simulation languages in Table~\ref{tab-8}. For most of the languages our method performs better than that of \newcite{duong2014can} and the three naive BiLSTM baselines. Directly training on projected data hurts the performance, which can be seen by comparing BiLSTM Projected and BiLSTM Ann+Proj. BiLSTM Annotated mostly outperforms MaxEnt Supervised, but both methods are worse than Duong et al.\@ and our BiLSTM Debias, which both use the projected data more effectively. The results show the debiasing layer makes more effective use of the projected data, improving the POS tagging accuracy.  

We show the learned bias transformation matrices for the different languages in Figure~\ref{fig-8}. The blue (dark) cells in the grids denote values that are most highly weighted. Note the strong diagonal, showing that the tags are mostly trusted, although there is also evidence of significant mass in off-diagonal entries. The worst case is in Greek (el) with many weak values on the diagonal. In this case, PRON and X appear to be confused for one another. The light cells are also important, showing tag combinations that the model learns to ignore, such as CONJ vs DET in Spanish (es) and PRON vs ADP in Swedish (sv). 
The tokens that are CONJ in Spanish (es) are seldom projected as DET. Overall, for most of languages the level of debiasing is modest, which might not come as a surprise given the large, clean parallel corpus for learning word alignments. 

\begin{table}
\begin{threeparttable}
\centering
\begin{tabular}{l c c}
\hline \bf \multirow{2}{*}{Model} & \multicolumn{2}{c}{\bf Accuracy}\\
 & {\small Malagasy}  & {\small Kinyarwanda} \\ \hline
BiLSTM Annotated  & 81.5 & 76.9\\
BiLSTM Projected &  67.2 & 61.9 \\
BiLSTM Ann+Proj &  78.6 & 73.2\\
\hline
MaxEnt Supervised & 80.0 & 76.4 \\
\newcite{duong2014can} & 85.3 & 78.3\\
BiLSTM Debias & 86.3 & 82.5\\
BiLSTM Debias {\small (Penn)} & 86.7 & 82.6\\
\newcite{garrette2013real} & 81.2 & 81.9\\
\hline
\end{tabular}
\caption{The POS tagging accuracy for various models in Malagasy and Kinyarwanda. The top results of the second part are taken from~\newcite{duong2014can}, evaluated on the same data split.}
\label{tab-mlg-kin}
\begin{tablenotes}\footnotesize
\item[*] Penn indicates the Penn treebank tagset. The proposed BiLSTM Debias can use different tagsets for the source language.
\end{tablenotes}
\end{threeparttable}
\end{table}

Now we present results for the two low-resource languages, Malagasy and Kinyarwanda, which both have much smaller parallel corpora. The results in Table~\ref{tab-mlg-kin} show that our method works better than all others in both languages, with a similar pattern of results as for the European languages. We also used the original Penn treebank tagset for both two languages. The results of BiLSTM Debias {\small(Penn)} show a small improvement, presumably due to the information loss in the mapping to the universal tagset. Note that our method outperforms the state of the art on both languages \cite{duong2014can,garrette2013real}.

To better understand the effect of the bias layer, we present the learned transformation matrices $A$ in Figure~\ref{fig-mlg-kin-nl}. Note the strong diagonal for Malagasy in Figure~\ref{fig-mlg-kin-nl}, showing that each tag is most likely to map to itself, however there are also many high magnitude off-diagonal elements. For instance nouns map to not just nouns, but also adjectives and numbers, but never pronouns (which are presumably well aligned). Comparing results of Malagasy and Kinyarwanda in Figure~\ref{fig-mlg-kin-nl}, we can see the divergence between the gold and projected tags is much greater in Kinyarwanda. This tallies with the performance results, in which we get stronger results and a greater improvement on Malagasy from using projection data where we had more parallel data.

\section{Conclusion}
In this paper we presented a technique for exploiting errorful cross-lingual projected annotations alongside a small amount of annotated data in the context of POS tagging. Projection on its own is unreliable and simple combination with gold is not sufficient to improve accuracy, even with only a tiny handful of gold annotations. To utilize both sources of data, we proposed a new model based on a bidirectional long short-term memory recurrent neural network, with a layer for explicitly handling projection labels. Over eight European and two real low-resource languages, our methods outperform other algorithms. Our technique is general, and is likely to prove useful for exploiting other noisy and biased annotations such as distant supervision and crowd-sourced annotations.

\section{Acknowledgments}
This work was sponsored by the Defense Advanced Research Projects Agency Information Innovation Office (I2O) under the Low Resource Languages for Emergent Incidents (LORELEI) program issued by DARPA/I2O under Contract No. HR0011-15-C-0114. The views expressed are those of the author and do not reflect the official policy or position of the Department of Defense or the U.S. Government. Trevor Cohn was supported by the Australian Research Council Future Fellowship (project number FT130101105).

\bibliography{acl2016}
\bibliographystyle{acl2016}

\end{document}